\documentclass[robotics,article,accept,moreauthors]{Definitions/mdpi}

\graphicspath{{./fig/}}

\DeclareMathOperator*{\argmax}{arg\,max}

\Title{Unified Promptable Panoptic Mapping with Dynamic Labeling Using Foundation Models}
\firstpage{1}
\pubvolume{1}
\issuenum{1}
\articlenumber{0}
\pubyear{2026}
\copyrightyear{2026}
\datereceived{27 November 2025}
\daterevised{14 January 2026}
\dateaccepted{22 January 2026}
\datepublished{26 January 2026}

\Author{{Mohamad} 
 Al Mdfaa $^{1}$\orcidA{}, Raghad Salameh $^{1}$, Geesara Kulathunga $^{2,}$*, Sergey Zagoruyko $^{3}$ and Gonzalo Ferrer $^{1,}$*}
\AuthorNames{Mohamad Al Mdfaa, Raghad Salameh, Geesara Kulathunga, Sergey Zagoruyko and Gonzalo Ferrer}

\address{%
$^{1}$\quad Applied AI Institute, {Moscow 121205}
, Russia; m.aimdfaa@applied-ai.ru (M.A.M.);\linebreak r.salameh@applied-ai.ru (R.S.)
\\
$^{2}$\quad School of Computer Science, University of Lincoln, {Lincoln LN6 7TS}
, UK\\
$^{3}$\quad Independent {Researcher, Dubai 500001, United Arab Emirates}
; {zagoruyko2@gmail.com}}

\corres{Correspondence: gkulathunga@lincoln.ac.uk (G.K.); g.ferrer@applied-ai.ru (G.F.)}

\abstract{Panoptic maps enable robots to reason about both geometry and semantics. However, open-vocabulary models repeatedly produce closely related labels that split panoptic entities and degrade volumetric consistency.
The proposed UPPM advances open-world scene understanding by leveraging foundation models to introduce a panoptic Dynamic Descriptor that reconciles open-vocabulary labels with unified category structure and geometric size priors. The fusion for such dynamic descriptors is performed within a multi-resolution multi-TSDF map using language-guided open-vocabulary panoptic segmentation and semantic retrieval, resulting in a persistent and promptable panoptic map without additional model training.
Based on our evaluation experiments, UPPM shows the best overall performance in terms of the map reconstruction accuracy and the panoptic segmentation quality. The ablation study investigates the contribution for each component of UPPM (custom NMS, blurry-frame filtering, and unified semantics) to the overall system performance. Consequently, UPPM preserves open-vocabulary interpretability while delivering strong geometric and panoptic accuracy{.}
}


\begin{document}

\section{Introduction}
Panoptic mapping connects 3D geometry with semantics so that robots can localize, plan, and~query in structured environments.
Systems such as SemanticFusion~\cite{mccormac2017semanticfusion}, \mbox{SLAM++ \cite{salas2013slam++}}, Panoptic Fusion~\cite{narita2019panopticfusion}, and DKB-SLAM~\cite{sun2025dkb} deliver spatial consistency but assume a predefined set of categories; open-vocabulary labels therefore fragment instances or fall back to generic labels.
Recent advances in high-capacity vision segmentation~\cite{li2023mask, zong2023detrs, zhu2024simple, zhou2025hypidecoder} and open-vocabulary or vision-language segmentation~\cite{zhang2023simple,li2024omg,ghiasi2022open,liu2023openseg,li2023ovseg} broaden the 2D label space, but~their predictions still live on individual images; they do not resolve how to reconcile those labels with volumetric priors, maintain cross-view consistency, or~reason about object scale in 3D.
{This gap motivates the core challenge addressed in this work: how to unify the rich, open-vocabulary labels produced by modern vision--language models into a consistent 3D panoptic representation without sacrificing either semantic expressiveness or \mbox{geometric accuracy}.}

A wave of recent works has extend open-vocabulary reasoning into 3D. Panoptic Vision--Language Feature Fields (PVLFF) learn a radiance field together with semantic and instance feature heads by distilling vision--language embeddings into a feature field and clustering the learned instance features~\cite{chen2024panopticvlf}. Despite achieving open-vocabulary panoptic segmentation, PVLFF must be optimized offline for each scene~\cite{chen2024panopticvlf}. OpenVox introduces a real-time, incremental open-vocabulary voxel representation that uses caption-enhanced detection and probabilistic instance voxels; its incremental fusion decomposes into instance association and live map evolution to increase robustness~\cite{openvox2024}. While effective, OpenVox focuses on instance association rather than unifying synonyms or transferring size~priors.

Other open-vocabulary mapping frameworks embed language features in 3D without panoptic constraints. ConceptFusion~\cite{jatavallabhula2023conceptfusion} projects pixel-aligned CLIP~\cite{radford2021learning} and DINO~\cite{caron2021emerging} features into 3D to enable multimodal queries; the resulting pixel-aligned embeddings capture fine-grained, long-tailed concepts but do not preserve instance continuity or temporal consistency. ConceptGraphs~\cite{gu2024conceptgraphs} associates class-agnostic masks across views to build an object-centric 3D scene graph and employs a large language model to caption each object and infer spatial relationships; this supports complex language queries but lacks volumetric geometry and per-voxel size priors. Clio~\cite{clio2024} formulates task-driven scene understanding through the information bottleneck, clustering 3D primitives into task-relevant objects and building a hierarchical scene graph, yet its segmentation granularity depends on task-specific thresholds. OpenScene~\cite{peng2023openscene} trains a 3D network to co-embed each point with text and image features in the CLIP space~\cite{radford2021learning}, enabling zero-shot queries of materials, affordances, and room types, but~its dense per-point features do not distinguish individual~instances.

\begin{figure}[H]
  
  \includegraphics[width=\linewidth]{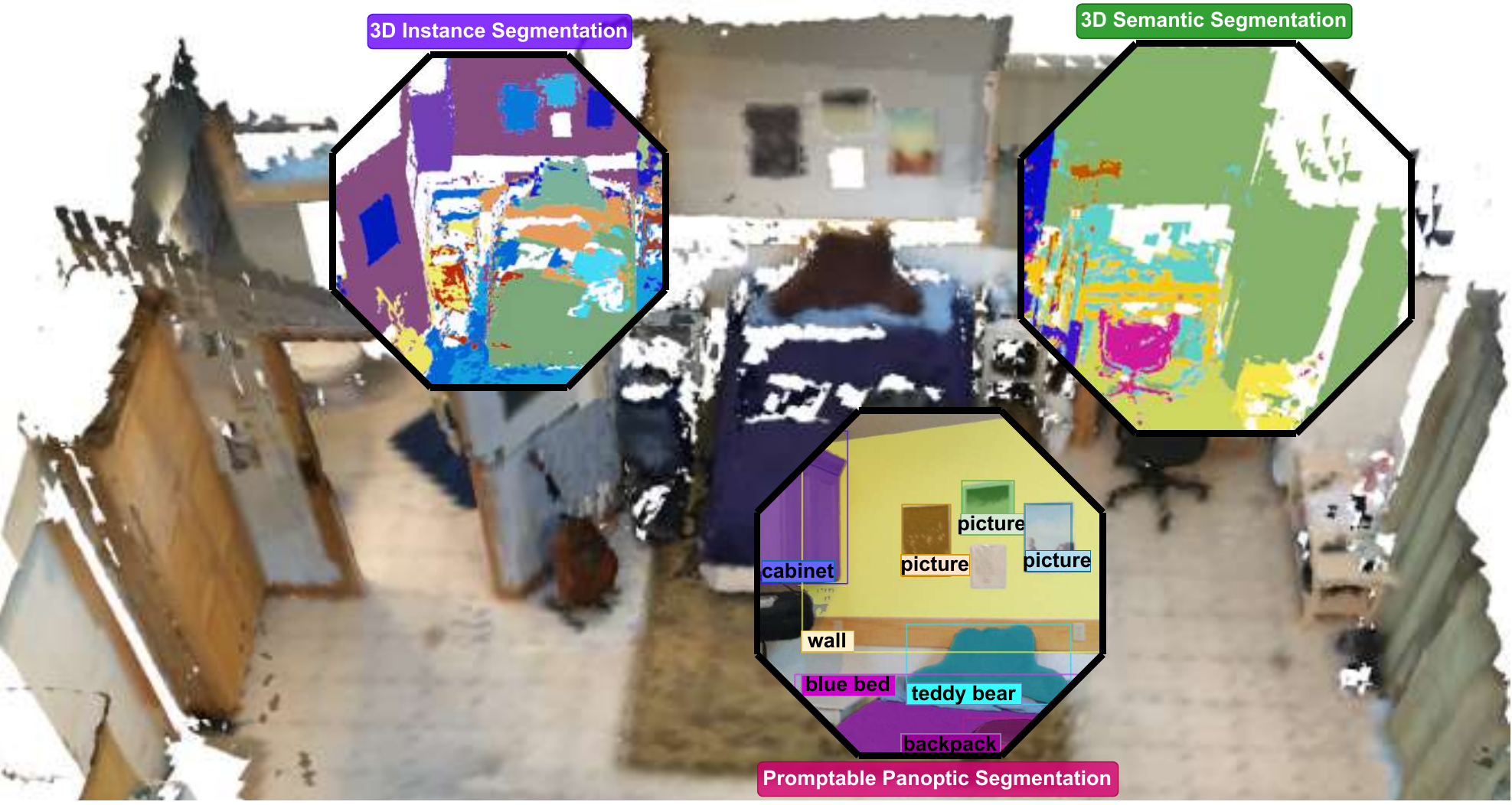}
  \caption{{Overview} 
 of Unified Promptable Panoptic Mapping (UPPM). Dynamic descriptors aggregate open-vocabulary labels, unified semantic categories, and~size priors and~are fused into a multi-resolution multi-TSDF map that preserves panoptic consistency while remaining queryable with natural-language~prompts.}
  \label{fig:teaser}
\end{figure}

\begin{figure}[H]

  \includegraphics[width=0.95\linewidth]{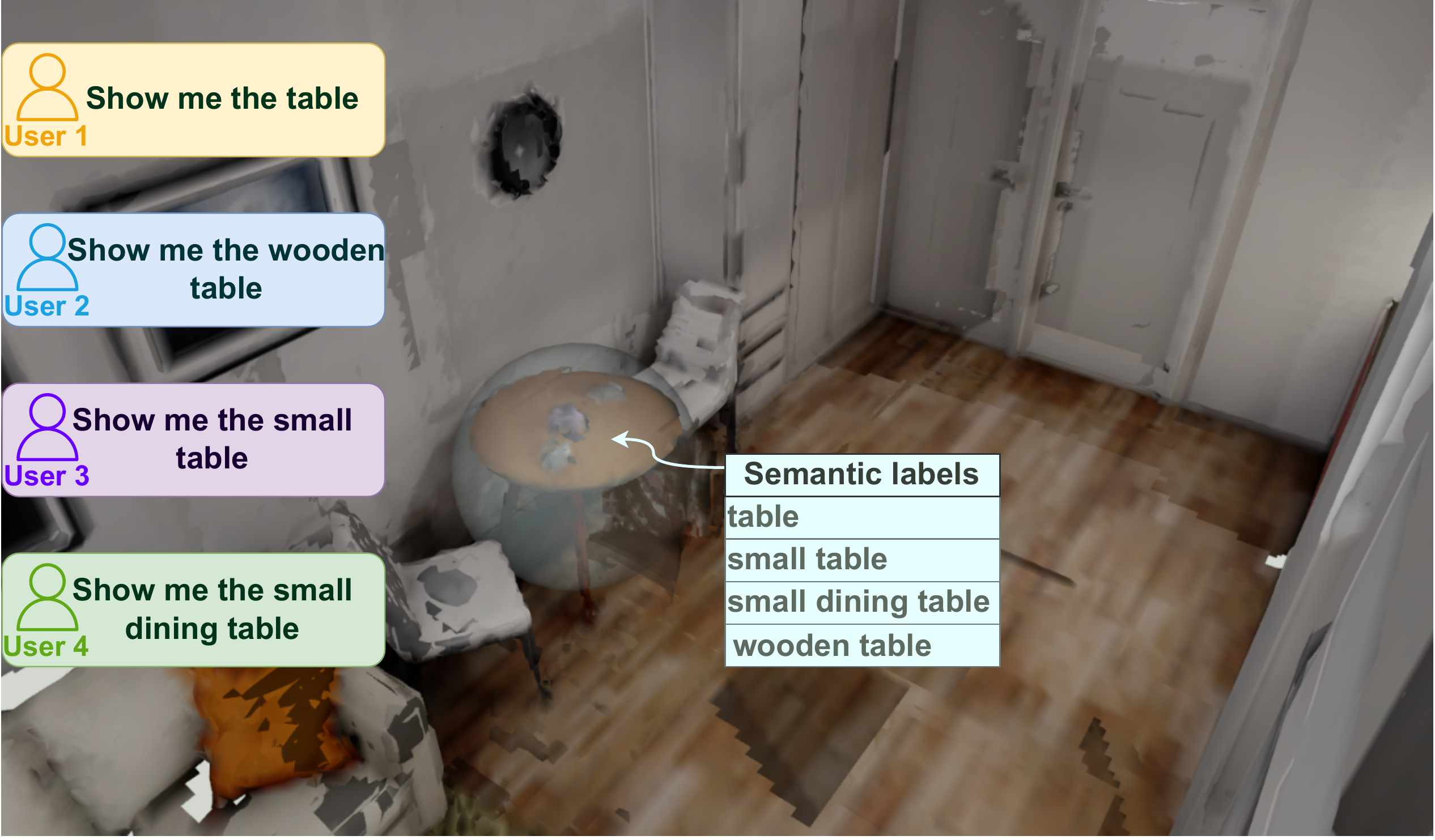}
  \caption{{Language}
-conditioned object retrieval via dynamic descriptors. Different textual queries, including synonyms and paraphrases, consistently retrieve the same 3D object because each submap stores all accumulated open-vocabulary labels alongside a unified category and size prior.}
  \label{fig:localization}
\end{figure}

{Unified Promptable Panoptic Mapping (UPPM) addresses these limitations through a novel dynamic descriptor mechanism that bridges the gap between open-vocabulary perception and consistent 3D panoptic mapping (\cref{fig:teaser}{). }}
{Unlike prior works that either require per-scene training (PVLFF~\cite{chen2024panopticvlf}) or focus solely on instance association without semantic unification (OpenVox~\cite{openvox2024}), UPPM introduces a fundamentally different approach: it aggregates open-vocabulary labels at the object level during fusion, producing persistent dynamic descriptors that maintain both semantic richness and categorical consistency.}
We use off-the-shelf vision--language models~\cite{huang2023tag2text} to generate elementary descriptors, where each object elementary descriptor conveys a rich open-vocabulary label, and then perform semantic retrieval to attach it to a unified category with an inherited volumetric size prior.
Rather than treating these elementary descriptors in isolation, UPPM groups them within the unified panoptic fusion block into a dynamic descriptor for each object.
Each dynamic descriptor stores the aggregated original captions, the~assigned unified category, and a size prior for a single object.
Each segmentation mask with its elementary descriptor are associated to the submap (i.e., 3D panoptic entity, which could be an object, a~piece of background, or free space) that has the highest Intersection over Union (IoU) between the predicted and rendered mask and the same class label.
To avoid spurious associations, a~minimum IoU of $\xi_{IoU} = 0.1$ is {required.} 
{This design represents a paradigm shift from detection-time labeling to fusion-time semantic aggregation, enabling UPPM to resolve the label fragmentation problem inherent in frame-by-frame open-vocabulary processing.}
UPPM therefore preserves the expressiveness of open-vocabulary prompts while producing category-consistent panoptic maps without per-scene training or graph-based reasoning.
The unified descriptors also enable promptable downstream queries, such as language-conditioned object retrieval, without~compromising geometric accuracy.
Figure~\ref{fig:localization} showcases how the resulting map enables language-conditioned retrieval without additional~training.

We assess UPPM on three evaluation regimes: Segmentation-to-Map evaluates the impact of 2D predictions on the quality of 3D reconstruction. Map-to-Map directly compares reconstructed 3D maps with ground-truth geometry, while Segmentation-to-Segmentation evaluates 2D panoptic segmentation~performance.



\vspace{-10pt}

This work has the following~contributions:
\begin{enumerate}
\item[(i)] {We introduce a panoptic dynamic descriptor mechanism that fundamentally changes how open-vocabulary labels are handled in 3D mapping.}
Rather than assigning labels per detection, UPPM aggregates the object open-vocabulary labels produced by vision--language models~\cite{huang2023tag2text} and post-processed by part-of-speech tagging~\cite{honnibal2013good}, maps them to a single unified semantic category and size prior, and~stores them as the dynamic descriptor (\cref{sec:dynamic_label_generation_and_segmentation,sec:sts}).
{This fusion-time aggregation---as opposed to detection-time labeling---is the key change that enables consistent panoptic maps from inconsistent open-vocabulary predictions.}
\item[(ii)] We build the UPPM pipeline around these dynamic descriptors to produce 3D panoptic maps that are geometrically accurate, semantically consistent, and~naturally queryable in language (\cref{sec:methods}).
{The pipeline operates without per-scene training, graph-based reasoning, or~neural radiance fields, making it suitable for real-time robotic applications.}
\item[(iii)] \textls[+15]{We extensively evaluate UPPM across Segmentation-to-Map, Map-to-Map, and} Segmentation-to-Segmentation regimes on three different datasets (ScanNet v2, RIO, and~Flat), and we~perform ablations of unified semantics, custom NMS, blurry-frame filtering, and~tag usage (\cref{sec:seg-to-map-eval,sec:map-to-map-eval,sec:seg-to-seg-eval,sec:unified-semantics,sec:nms,sec:blurry-frames-filtering}).
Together, these experiments show that dynamic descriptors improve both geometric reconstruction and panoptic quality while enabling downstream language-conditioned tasks such as object retrieval.
\end{enumerate}

\section{Related~Work}
\label{sec:related_work}

Recent semantic mapping systems span multi-resolution volumetric fusion and metric-semantic SLAM. Panoptic Multi-TSDFs~\cite{schmid2022panoptic} and the hierarchical DHP-Mapping framework~\cite{hu2024dhp} maintain panoptic submaps with label optimization, while Fusion++ \cite{mccormac2018fusion++} instantiates object-level volumetric models and Kimera~\cite{rosinol2020kimera} produces metric-semantic reconstructions and dynamic scene graphs. Likewise, methods such as~\cite{mahmoud2022improved} explore semantic segmentation and room-layout cues to improve robustness. Despite strong performance, these pipelines typically operate with closed vocabularies, {which limits their applicability in the real-world scenarios}.

{\subsection{Panoptic 3D Scene~Understanding}}
{A growing body of work addresses panoptic understanding of 3D scenes from various input modalities.
Panoptic NeRF~\cite{fu2022panopticnerf} performs 3D-to-2D label transfer for urban scene segmentation by lifting 2D panoptic labels into neural radiance fields, demonstrating effective cross-view consistency for outdoor environments.
Dahnert~et~al.~\cite{dahnert2021panoptic3d} reconstruct panoptic 3D scenes from a single RGB image by jointly predicting geometry and semantics, though~the single-view setting limits geometric accuracy compared to multi-view fusion.
PanopticDepth~\cite{gao2022panopticdepth} unifies depth estimation with panoptic segmentation in a single framework, showing that joint reasoning improves both tasks; however, it operates on individual images without temporal fusion.
For video understanding, Yang~et~al.~\cite{yang2023panopticvsg} introduce panoptic video scene graph generation that captures spatio-temporal relations between objects, extending panoptic reasoning to the temporal domain.
PanopticPartFormer++ \cite{li2024panopticpartformerpp} provides a unified and decoupled architecture for panoptic part segmentation, achieving strong results on part-level understanding but focusing on 2D rather than 3D reconstruction.
Panoptic Lifting~\cite{siddiqui2023panopticlifting} lifts 2D panoptic predictions into neural fields for 3D scene understanding, enabling open-vocabulary queries but requiring per-scene optimization.
Unlike these approaches, UPPM operates incrementally without per-scene training and explicitly unifies open-vocabulary labels into consistent dynamic descriptors during \mbox{volumetric fusion}.}

\subsection{Open-Vocabulary 3D~Mapping}
To relax vocabulary constraints, vision--language extensions inject language supervision into volumetric pipelines. PVLFF~\cite{chen2024panopticvlf} learns a radiance field with semantic and instance feature fields and performs open-vocabulary panoptic segmentation by clustering instance features; however, it requires per-scene offline training and is not designed for real-time incremental mapping. OpenVox~\cite{openvox2024} introduces an incremental instance-level open-vocabulary mapping framework with caption-enhanced detection and a probabilistic voxel representation. It maintains an embedding codebook for each instance and emphasizes robust association and mapping, rather than unifying synonyms into a common category or transferring volumetric size priors. In~contrast, UPPM leverages off-the-shelf foundation models without additional training, adds a semantic retrieval stage to assign each object a unified category and size, and~fuses dynamic descriptors into a multi-resolution multi-TSDF~map.

A number of recent works further expand open-vocabulary mapping by exploring new architectures and representations. FindAnything~\cite{laina2025findanything} integrates vision--language features into volumetric occupancy submaps to support object-centric mapping with natural-language queries on lightweight platforms. OVO~\cite{martins2025ovo} employs online detection and tracking of 3D segments with CLIP-based descriptor fusion to achieve open-vocabulary mapping within a SLAM. RAZER~\cite{patel2025razer} combines GPU-accelerated TSDF reconstruction with hierarchical association and spatio-temporal aggregation to produce zero-shot panoptic maps, reporting real-time performance on modern GPUs. OpenGS-Fusion~\cite{yang2025opengsfusion} couples 3D Gaussian splats with TSDFs and introduces language-guided thresholding to improve volumetric segmentation quality. DualMap~\cite{jiang2025dualmap} uses a dual representation (abstract/global and concrete/local) and a hybrid segmentation front end to handle dynamic scenes and language-conditioned navigation. MR-COGraphs~\cite{gu2025mrcographs} compresses semantic features in scene graphs for communication-efficient multi-robot mapping. 3D-AVS~\cite{wei2025avs} proposes automatic vocabulary discovery for LiDAR point clouds, and~Open-Vocabulary Functional 3D Scene Graphs~\cite{zhang2025openfungraph} encode objects, interactive elements, and~functional relations using vision--language and large language models. These methods illustrate the community's pivot toward open-vocabulary and hierarchical understanding, but~many rely on neural radiance fields, Gaussian splats, or~graph reasoning and do not unify synonyms or transfer volumetric size priors during~mapping.

\subsection{Open-Set Scene Graphs and Feature~Maps}
OpenScene~\cite{peng2023openscene} computes CLIP-aligned features for every 3D point using multi-view 2D fusion and 3D distillation to enable task-agnostic open-vocabulary retrieval; it constructs a semantic map but does not perform instance-level segmentation. Beyond~OpenScene, scene-graph and radiance-field approaches such as Clio~\cite{clio2024}, Concept-Graphs~\cite{gu2024conceptgraphs}, LangSplat~\cite{langsplat2024}, Bayesian-Fields~\cite{bayesianfields2023}, and~OpenGaussian~\cite{opengaussian2024} enable language-guided reasoning, yet they often trade off panoptic completeness or require per-query optimization. HOV-SG~\cite{hovsg2024} builds hierarchical open-vocabulary 3D scene graphs by extracting CLIP embeddings per SAM mask and back-projecting fused features to 3D; unlike UPPM, it does not normalize embeddings into a fixed unified category space with inherited size priors, so it does not directly yield category-consistent panoptic instance maps. Consistent with semantics-conditioned scale choices in panoptic mapping~\cite{schmid2022panoptic,mccormac2018fusion++}, UPPM maintains a coarse size prior (Small/Medium/Large) during~mapping.

\subsection{Foundation Vision--Language~Models}
Foundation models~\cite{awais2025foundation} provide the building blocks for open-world perception. Multimodal encoders such as CLIP~\cite{radford2021learning} supply open-vocabulary embeddings; promptable segmentation models like Segment Anything~\cite{kirillov2023segment} yield category-agnostic masks; and multimodal assistants including LLaVA~\cite{liu2023visual} generate rich scene descriptions that can seed dynamic descriptors.
{Beyond discriminative models, diffusion-based foundation approaches have emerged as powerful generative priors.
For instance, EDiffSR~\cite{xiao2024ediffsr} demonstrates how diffusion probabilistic models can enhance feature representations through iterative refinement, a~principle that could potentially be adapted for semantic feature enhancement in mapping pipelines.
While diffusion models have primarily been explored for image generation and super-resolution, their ability to model complex distributions suggests future opportunities for improving label consistency in open-vocabulary mapping.
Recent work on pseudo-label guided optimization~\cite{jin2026pseudolabel} has shown that iterative refinement of noisy predictions from foundation models can improve downstream segmentation quality, a~strategy complementary to UPPM's semantic retrieval approach.}

ConceptFusion~\cite{jatavallabhula2023conceptfusion}, OVIR-3D~\cite{pmlr-v229-lu23a}, and~OpenMask3D~\cite{takmaz2023openmask3d} handle open-vocabulary cues in 3D but lack a shared categorical structure. UPPM combines dynamic descriptors with unified semantics so that the map remains panoptic while supporting promptable queries and coarse-to-fine relational reasoning within a single volumetric~representation.

\unskip

\section{Method}
\label{sec:methods}
\begin{figure}[H]

 \includegraphics[width=0.98\linewidth]{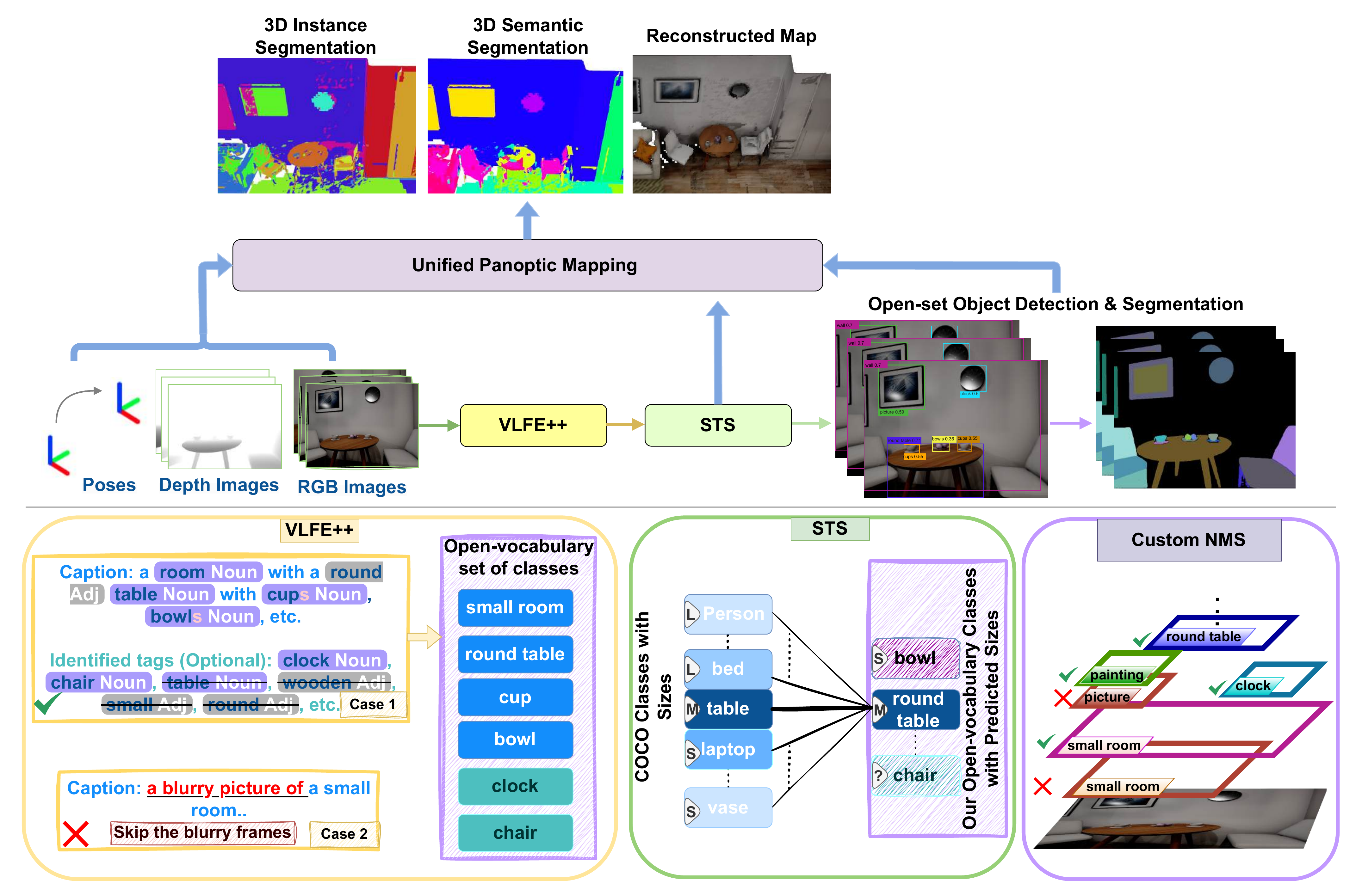}

 \caption{{System} 
 {overview of Unified Promptable Panoptic Mapping (UPPM)}.
 RGB frames are processed by VLFE++ to generate caption- and tag-based open-vocabulary labels. Semantic Retrieval (SR) maps these labels to unified semantic categories and size priors, forming elementary descriptors $e_i = \langle o_i, E(o_i), \hat{c}_i, \hat{s}_i \rangle$. Open-vocabulary promptable panoptic segmentation consume these descriptors, together with custom NMS, to~produce semantic and instance segmentations. Unified panoptic fusion then fuses geometry and elementary descriptors into a multi-resolution multi-TSDF representation with one dynamic descriptor per object, producing the unified promptable panoptic map (UPPM) $\mathcal{M} = (\mathcal{M}_g \bigoplus \mathcal{M}_d)$.}
 \label{fig:system-overview}
\end{figure}

\textls[+15]{UPPM addresses the visual semantic mapping problem by utilizing foundation} \mbox{models~\cite{kirillov2023segment,liu2023grounding,song2020mpnet,huang2023tag2text, zhang2024recognize}} to construct dynamic descriptors that help to build precise reconstructions with rich semantics. The~term `dynamic descriptor' here refers to the object-level data structure formed by aggregating all open-vocabulary cues observed for the same object over time, a~shared unified category and size prior, an~aggregation that is performed during the unified panoptic fusion stage.
The mapping pipeline follows a systematic multi-stage process: 
First, with~the help of vision--language models, we extract rich open-vocabulary labels from the RGB inputs.
Second, a~semantic retrieval module attaches to these labels a unified category with the associated volumetric size priors, yielding elementary descriptors for each label.
Third, language-conditioned open-vocabulary object detection and promptable image segmentation are guided by these elementary descriptors.
Finally, during~the unified panoptic fusion stage, the elementary descriptors associated with each object are aggregated into a single dynamic descriptor. Consequently, the~resulting semantic and instance segmentations with the dynamic descriptors are fused into a unified promptable panoptic map (UPPM) that is both geometrically accurate and semantically consistent. The~complete mapping pipeline is shown in \cref{fig:system-overview}.

\subsection{Problem~Formulation}
Given a sequence of posed RGBD images $\mathcal{U} = \{U_1, \ldots, U_T\}$ where each\linebreak   $U_t = (I_t, D_t, P_t)$ consists of an RGB image $I_t \in \mathbb{R}^{H \times W \times 3}$, a~depth image $D_t \in \mathbb{R}^{H \times W}$, and~a camera pose $P_t \in SE(3)$, our goal is to construct a unified promptable panoptic map $\mathcal{M}$ that satisfies the following objectives:

\begin{enumerate}
  \item {Geometric Reconstruction}
: Generate an accurate 3D reconstruction of the environment represented as a multi-resolution multi-TSDF map $\mathcal{M}_g$.
  
  \item {Dynamic Labeling with Unified Semantics:} Construct the set of dynamic descriptors $\mathcal{M}_d$ by first generating elementary descriptors from open-vocabulary cues and then, during~the unified panoptic fusion stage, aggregating all elementary descriptors that refer to the same object across frames into a single dynamic descriptor.
\end{enumerate}
{Specifically}
, we assign a dynamic descriptor $d_i \in \mathcal{M}_d$ to each unique object $o_i \in \mathcal{O}$, where $\mathcal{O}$ is the set of all objects in the scene. Each dynamic descriptor collects all open-vocabulary labels that have been associated with that object across frames and consolidates them under a single unified category and size prior. Formally, $d_i = \langle O_i, \hat{c}_i, \hat{s}_i \rangle$, where $O_i$ is the set of open-vocabulary labels aggregated for object $o_i$, $\hat{c}_i \in \mathcal{C}$ is the unified category chosen through \eqref{eq:unified-category}, and~$\hat{s}_i$ is the inherited size prior. We choose the COCO-Stuff classes~\cite{caesar2018coco} as the fixed unified categories due to their comprehensive coverage of common indoor objects and well-established hierarchy, providing a robust foundation for semantic mapping tasks.
The final output is a unified promptable panoptic map $\mathcal{M} = (\mathcal{M}_g \bigoplus \mathcal{M}_d)$, where the fusion operator $\bigoplus$ combines geometric reconstruction $\mathcal{M}_g$ with dynamic descriptors $\mathcal{M}_d$ into a unified volumetric representation integrating spatial and semantic~information.

\unskip

\subsection{Dynamic Labeling and~Segmentation}
\label{sec:dynamic_label_generation_and_segmentation}
As shown in \cref{fig:system-overview}, the~per-frame processing pipeline consists of three modules that construct structured dynamic descriptors from RGB images. Visual--Linguistic Feature Extraction++ (VLFE++) generates open-vocabulary labels, Semantic Retrieval (SR) maps such labels to unified categories, and~the open-vocabulary panoptic segmentation module consumes these labels to produce semantic and instance segmentations that are ready for~fusion.

\subsubsection{Visual--Linguistic Feature Extraction++ (VLFE++)}
\label{sec:vlfepp}
VLFE++ leverages vision--language models to gather open-vocabulary labels from visual input.
{In this work, ``open-vocabulary'' refers to the ability to recognize and label objects using freely generated textual descriptions from vision--language models, including novel object categories not present in fixed training vocabularies (e.g., ``vintage typewriter'', ``ceramic vase with floral pattern'') as well as descriptive phrases generated by the \mbox{captioning model}.}

As shown in \cref{fig:system-overview}, VLFE++ applies a three-step extraction procedure:

\begin{itemize}
\item {Visual Linguistic Feature Extraction (VLFE):}
{A pretrained vision--language model, Tag2Text} \cite{huang2023tag2text}{, is employed to produce both image captions and identified tags from a single forward pass.}
{Captions provide contextual descriptions (e.g., ``a wooden desk with a laptop and books''), while tags provide object-level labels (e.g., ``wooden desk'', ``laptop'', ``book'', ``chair'').}
{Captions and tags undergo sentence parsing, as explained in the next step.}
{As shown in} \cref{fig:system-overview}, {the two streams are then concatenated into a unified candidate label set, where caption-derived labels retain a provenance flag for downstream prioritization in NMS} (\cref{sec:ovps}).

\item {Part-Of-Speech (POS) Tagging:} A single-layer perceptron network~\cite{honnibal2013good} is applied to extract nouns, noun phrases, and~modifiers. The~resulting candidate list is passed to the next~step.

\item {Lemmatization:} Lemmatization~\cite{fellbaum2010wordnet} is used to map the derived forms from the candidate list to their normalized base forms (e.g., `apples' $\rightarrow$ `apple'), providing canonical labels for {semantic} retrieval.
{This normalization ensures that morphological variants (``chairs'' vs. ``chair'') map to the same unified category during semantic retrieval.}
\end{itemize}

The generated open-vocabulary labels serve as input for the next pipeline component, {Semantic Retrieval (SR)}.
{Note that Tag2Text also provides a confidence indicator for image quality; frames flagged as blurry or low-quality are optionally filtered before processing (\cref{sec:blurry-frames-filtering}).}

\subsubsection{Semantic Retrieval (SR)}
\label{sec:sts}

SR maps open-vocabulary  labels to unified categories while preserving these labels.
Let $\mathcal{O}_{vocab}$ be the set of open-vocabulary labels from VLFE++, and let $\mathcal{C}$ be the set of fixed unified categories, where each class $(c_i \in \mathcal{C})$ has a corresponding voxel size attribute $(\nu_i)$, which takes one of the following empirical scale priors: small ($\nu=2$ cm), medium ($\nu=3$~cm), or~large ($\nu=5$ cm), which were introduced in PanMap~\cite{schmid2022panoptic}. This choice maintains compatibility with established panoptic mapping pipelines and ensures consistent comparison. We also keep $\nu_{freespace}=30$ cm for all settings. Our goal is to determine the most appropriate unified category $\hat{c}$ for any given open-vocabulary label $o \in \mathcal{O}_{vocab}$:
\begin{equation}
\hat{c} = \argmax_{c \in \mathcal{C}} \cos(E(o), E(c)),
\label{eq:unified-category}
\end{equation}
where $\cos(\cdot)$ is the cosine similarity between embeddings and $E(\cdot)$ denotes the embedding function. The~predicted size $\hat{\nu} = \nu_{\hat{c}}$ is inherited from the assigned unified category $\hat{c}$. Following Panoptic Multi-TSDFs~\cite{schmid2022panoptic}, each unified category carries a manually curated volumetric prior; UPPM transfers this size to new elementary descriptors through SR so that dynamic descriptors remain metrically grounded. We employ MPNet~\cite{song2020mpnet} for embedding generation and perform semantic search~\cite{johnson2019billion} using cosine similarity to select the closest unified category.
For each open-vocabulary label $o_i \in \mathcal{O}_\text{vocab}$, SR finds the corresponding unified category $\hat{c}$ and predicts its size $\hat{\nu}$. These assignments yield an elementary descriptor $e_i = \langle o_i, \hat{c}_i, \hat{\nu}_i \rangle$ for the label. During~the unified panoptic fusion stage, these elementary descriptors are aggregated for each object into a single dynamic descriptor that collects all of its labels and shares a unified category and size prior. The~elementary descriptor for a label is defined as follows:
\begin{equation}
e_i = \langle o_i, \hat{c}_i, \hat{\nu}_i \rangle,
\end{equation}
where $o_i$ is the open-vocabulary label from VLFE++, $\hat{c}_i$ is the unified category assigned by \eqref{eq:unified-category}, and~$\hat{\nu}_i$ is the inherited size prior. These elementary descriptors drive the open-vocabulary promptable panoptic segmentation and later serve as building blocks for the object-level dynamic descriptors formed in the unified panoptic fusion stage (\cref{fig:localization}).

\subsubsection{Open-Vocabulary Panoptic~Segmentation}
\label{sec:ovps}

UPPM employs Grounding-DINO~\cite{liu2023grounding} as the object detector, leveraging its capability to process curated open-vocabulary labels. Grounding-DINO generates precise bounding boxes that serve as input prompts for the Segment Anything Model (SAM) \cite{kirillov2023segment}, enabling the creation of high-quality segmentation for each detected object in the scene.
Furthermore, as~shown {in} 
\cref{sec:nms}, a~custom Non-Maximum Suppression (NMS) technique is proposed to address the open-vocabulary detection challenges~\cite{liu2023grounding}. While traditional NMS approaches like per-class NMS~\cite{girshick2014rich} remove redundant bounding boxes based on Intersection-over-Union (IoU) and confidence scores within the same class, our custom NMS {uses a coordinate-based proximity criterion to detect duplicates:}

\begin{enumerate}
\item {{Near-duplicate detection:} Two bounding boxes are considered duplicates if their corresponding corners (top-left and bottom-right) are within a spatial threshold $\tau_{coord}$.}
{This coordinate-based criterion is more precise than IoU for detecting near-identical detections that arise from similar open-vocabulary prompts (e.g., ``chair'' and ``wooden chair'' detecting the same object).}
{When duplicates are detected, the~box with the lower confidence score is suppressed.}

\item {{Containment-based suppression}: If one bounding box is fully contained within another and both share the same label, the~contained box is suppressed.}
{This handles cases where a detection at multiple scales produces nested boxes for the same object.}
\end{enumerate}

{The coordinate threshold $\tau_{coord} = 1.5$ pixels was empirically determined to capture near-duplicate detections while preserving distinct nearby objects.}
{This approach reduces redundant associations by 22.2\% (recall improvement from 63.64\% to 100\%), as shown in} \cref{tab:custom-nms-vs-traditional-nms}, {because coordinate-based matching precisely identifies duplicate detections from semantically similar prompts.}
{For Grounding-DINO, the~box threshold is set to 0.35 and the text threshold to 0.25; SAM uses its default prompt-based segmentation settings.}

\begin{table}[H]
\small

  \caption{Comparison of traditional NMS~\cite{girshick2014rich} and our custom NMS on the Flat dataset. Custom NMS demonstrates better performance across all~metrics.}
     \setlength{\tabcolsep}{15pt}
  \begin{tabularx}{\textwidth}{lccc}
    \toprule
    \textbf{Method} & \textbf{Precision [\%] (\boldmath{$\uparrow$})} & \textbf{Recall [\%] (\boldmath{$\uparrow$})} &
    \textbf{F1-Score [\%] (\boldmath{$\uparrow$})}\\
    \midrule
    NMS~\cite{girshick2014rich} & 91.3 & 63.64 & 75.0 \\
    Custom NMS (Ours) & {94.6} & {100.0} & {97.22} \\
    \bottomrule
  \end{tabularx}
  \label{tab:custom-nms-vs-traditional-nms}
\end{table}

\subsection{Unified Panoptic Fusion (UPF)}
Building upon the open-vocabulary panoptic segmentation results, we integrate these outputs into a comprehensive 3D representation by adopting a panoptic multi-TSDFs-based approach~\cite{schmid2022panoptic}. Traditional panoptic mapping approaches rely on fixed semantic categories, limiting their ability to handle novel objects. Our Unified Panoptic Fusion extends beyond these limitations by integrating dynamic descriptors while maintaining semantic consistency through unified categorization.
We build upon the object-centric mapping framework introduced in~\cite{schmid2022panoptic}, specifically leveraging its submap-based architecture. Submaps are localized volumetric representations that divide large-scale environments into smaller, manageable parts, where each submap contains geometric and semantic data including panoptic information, object instances, and~semantic categories, along with transformation and tracking data. These submaps enable efficient processing of large-scale environments while significantly reducing computational complexity.
UPF treats each object, background region, or~free space as its own submap stored in a separate TSDF grid, with~its own instance identity and semantic class. The~rendered masks are associated with panoptic segmentations based on Intersection-over-Union (IoU). The~new observation is fused into the submap only if the IoU exceeds a small threshold; otherwise, a~new submap is created. If~$\mu_e$ denotes the mask for the elementary descriptor $e$ in the new frame, and~$\rho_s$ denotes the rendered mask of a submap $s$, then the assigned submap $s^{*}$ is as shown in \cref{eq:submap_association}.
\begin{equation}
s^{*}(e) = \begin{cases}
\arg\max\limits_{s \in \mathcal{S}} \mathrm{IoU}(\mu_e, \rho_s), & \text{if } \max_{s \in \mathcal{S}} \mathrm{IoU}(\mu_e, \rho_s) \geq \xi_{\mathrm{IoU}},\\
s_{new}, & \text{otherwise},
\end{cases}
\label{eq:submap_association}
\end{equation}
where $\mathcal{S}$ is the set of submaps, $s_{new}$ is a new submap, and $\xi_{\mathrm{IoU}}$ is a small association threshold.

\begin{figure}[H]

  \includegraphics[width=0.95\linewidth]{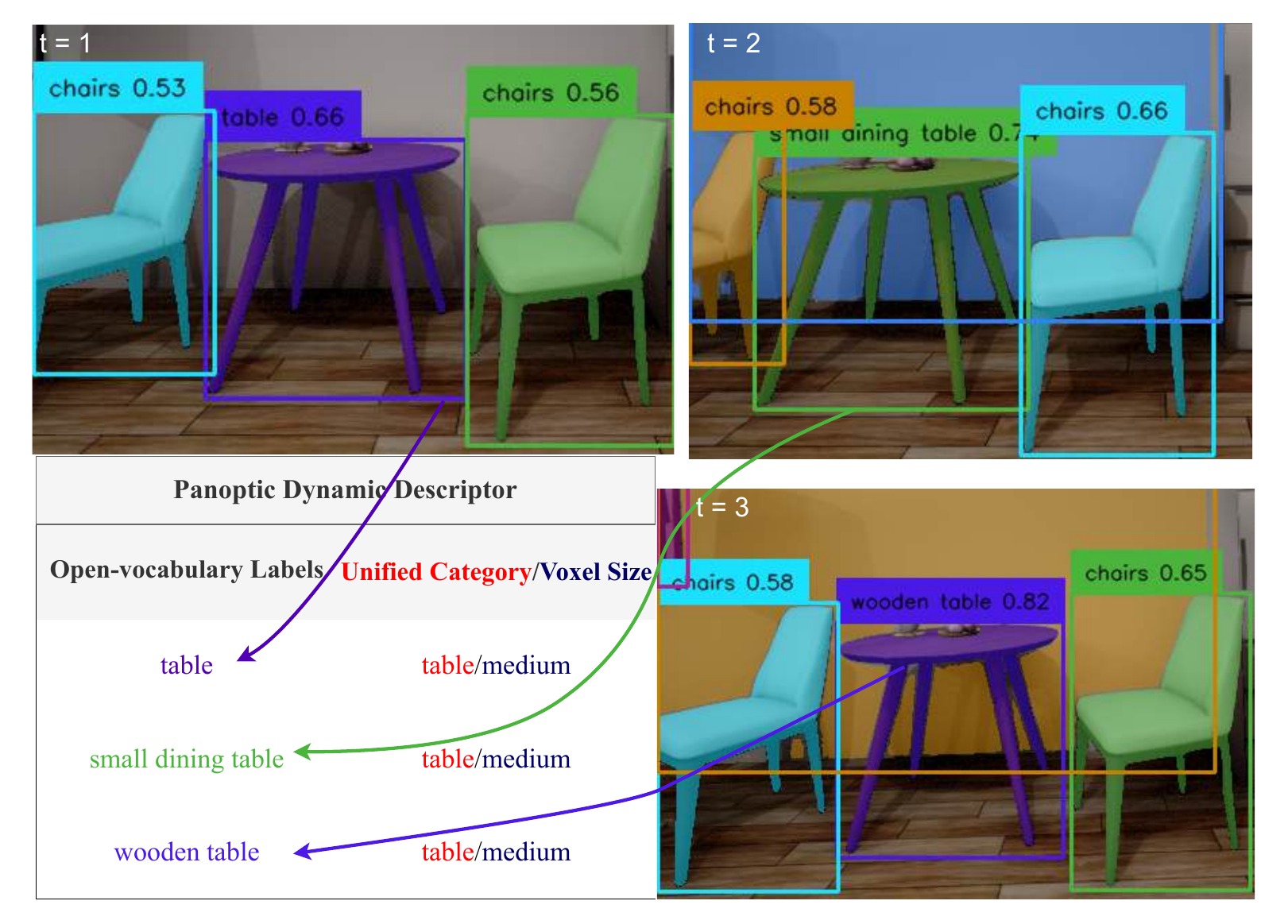}
  \caption{{Panoptic} 
 dynamic descriptors unify open-vocabulary labels over time. As~observations accumulate across frames, the~descriptor remains attached to the same object in the panoptic map, preserving promptable yet category-consistent~semantics.}
  \label{fig:unified_concept}
\end{figure}

\textls[-15]{Our key contribution is to enrich these submaps with dynamic descriptors, as~illustrated in \cref{fig:unified_concept}. Whereas PanMap~\cite{schmid2022panoptic}} assigns a fixed semantic label to every submap, UPPM augments each submap with a dynamic descriptor that aggregates all open-vocabulary labels observed for that object across frames, along with the retrieved unified category and inherited voxel size prior. As~new frames are fused, their elementary descriptors are merged into the submap so that closely related elementary descriptors accumulate into a single consolidated dynamic descriptor. These descriptors travel with the submap through the association and change-detection machinery described above, ensuring that there is exactly one descriptor per object rather than per detection. This grouping enables flexible open-vocabulary querying without sacrificing the hard panoptic consistency afforded by the underlying PanMap~framework.

\vspace{-3pt}

{{Dynamic Descriptor Aggregation.}}
{When a new elementary descriptor is associated with an existing submap,} \cref{eq:submap_association}{, the~aggregation proceeds as follows:}
\begin{enumerate}
\item {{Label accumulation:} The open-vocabulary label is appended to the submap's \mbox{label set.}}
\item {{Unified category resolution:} If the incoming unified category matches the submap's existing category, no change is made.}
{If they differ, we apply majority voting across all accumulated labels.}

\item {{Size prior inheritance:} As explained in the previous step, the~size prior is inherited from the unified category, ensuring consistent voxel resolution throughout the \mbox{object's lifetime.}}
\end{enumerate}
{{This} aggregation strategy ensures temporal consistency: even if an object such as a ``table'' is labeled as ``table'' in one frame and ``dining table'' in another---two distinct categories in COCO-Stuff---majority voting consolidates them into a single unified category. Meanwhile, open-vocabulary labels continue to accumulate while the unified category (\mbox{e.g., COCO-Stuff} ``table'') remains stable.}
{The accumulated labels enable flexible downstream queries---a user can search for ``kitchen table'', ``eating table'', or~``dining table'' and retrieve the same object.}
The Unified Panoptic Fusion component takes as input posed RGBD data, dynamic descriptors, detected objects, and~panoptic segmentation, reconstructing the Unified Promptable Panoptic Map (UPPM) $\mathcal{M}$ by integrating this information into the enhanced submap structure. The~resulting map is accurately reconstructed and enriched with dynamic descriptors, facilitating downstream tasks such as navigation and~localization.

\section{Results}
\label{sec:results}
We evaluate UPPM across three regimes:
{Segmentation-to-Map} measures how 2D panoptic segmentations drive 3D reconstruction when PanMap~\cite{schmid2022panoptic} is paired with different segmentation backbones across ScanNet v2~\cite{dai2017scannet}, RIO~\cite{wald2019rio}, and~Flat~\cite{schmid2022panoptic}. {Geometric metrics include accuracy (cm), completeness (cm), Chamfer distance (cm), Hausdorff distance (cm), completion ratio (\%), and~F1-score; runtime is reported in frames per second (FPS).}
{Map-to-Map} compares reconstructed maps directly against the ground-truth geometry using {accuracy (cm), completeness (cm), and~Chamfer-L1 (cm)}.
{Segmentation-to-Segmentation} evaluates panoptic segmentation quality on Flat using {Panoptic Quality (PQ), Recognition Quality (RQ), Segmentation Quality (SQ), and~mAP at IoU thresholds of 0.3, 0.4, and~0.5}.

\subsection{Segmentation-to-Map~Evaluation}
\label{sec:seg-to-map-eval}

The impact of segmentation quality on geometric reconstruction is assessed by comparing UPPM with PanMap~\cite{schmid2022panoptic}, each deployed with different segmentation backbones: MaskDINO~\cite{li2023mask}, OpenSeeD~\cite{zhang2023simple}, OMG-Seg~\cite{li2024omg}, and~ground-truth segmentation (GTS).

UPPM achieves strong performance across multiple metrics (\cref{tab:combined-eval}). The~proposed method attains optimal accuracy on RIO ($1.55$ {cm}
) and Flat ($0.61$ cm) with competitive performance on ScanNet v2 ($3.30$ cm), where {Accuracy} measures how close reconstructed points are to ground truth, calculated as mean absolute error from reconstruction ($R$) to ground truth ($G$). This improved accuracy on RIO shows UPPM's robustness to motion blur and sensor noise common in real-world datasets.

\begin{table}[H]
\scriptsize
\setlength{\tabcolsep}{2.67pt}
  \renewcommand{\arraystretch}{1.2}
\renewcommand{\aboverulesep}{.1pt}
\renewcommand{\belowrulesep}{.1pt}
  \caption{{Quantitative} 
 evaluation across all datasets. All baselines use the PanMap framework~\cite{schmid2022panoptic} with different segmentation backbones. Abbreviations: Comp. = Completion, Acc. = Accuracy, Chamf. = Chamfer distance, Haus. = Hausdorff distance, C.R. = Completion Ratio, F1 = F1-Score, FPS = Frames Per Second. The~highlight colors indicate the ranking for each metric: \textcolor[RGB]{181,229,80}{best}, \textcolor[RGB]{75,139,190}{second best}, and~\textcolor[RGB]{230,92,90}{third best}. {$\downarrow$ indicates lower is better; $\uparrow$ indicates higher is better.}}

\begin{adjustwidth}{-\extralength}{0cm}
\begin{tabularx}{\fulllength}{lccccccccccccccccc}

    \toprule
    & \multicolumn{5}{c}{\textbf{ScanNet v2}} & \multicolumn{5}{c}{\textbf{RIO}} & \multicolumn{7}{c}{\textbf{Flat}} \\
    \cmidrule{2-6} \cmidrule{7-11} \cmidrule{12-18}
  \textbf{Method} & \shortstack[c]{\textbf{Comp.}\\\textbf{[cm] (\boldmath{$\downarrow$})}} & \shortstack[c]{\textbf{Acc.}\\\textbf{[cm] (\boldmath{$\downarrow$})}} & \shortstack[c]{\textbf{Chamf.}\\\textbf{[cm] (\boldmath{$\downarrow$})}} & \shortstack[c]{\textbf{Haus.}\\\textbf{[cm] (\boldmath{$\downarrow$})}} & \shortstack[c]{\textbf{C.R.}\\\textbf{[\%](\boldmath{$\uparrow$})}} & \shortstack[c]{\textbf{Comp.}\\\textbf{[cm] (\boldmath{$\downarrow$})}} & \shortstack[c]{\textbf{Acc.}\\\textbf{[cm] (\boldmath{$\downarrow$})}} & \shortstack[c]{\textbf{Chamf.}\\\textbf{[cm] (\boldmath{$\downarrow$})}} & \shortstack[c]{\textbf{Haus.}\\\textbf{[cm] (\boldmath{$\downarrow$})}} & \shortstack[c]{\textbf{C.R.}\\\textbf{[\%](\boldmath{$\uparrow$})}} & \shortstack[c]{\textbf{Comp.}\\\textbf{[cm] (\boldmath{$\downarrow$})}} & \shortstack[c]{\textbf{Acc.}\\\textbf{[cm] (\boldmath{$\downarrow$})}} & \shortstack[c]{\textbf{Chamf.}\\\textbf{[cm] (\boldmath{$\downarrow$})}} & \shortstack[c]{\textbf{Haus.}\\\textbf{[cm] (\boldmath{$\downarrow$})}} & \shortstack[c]{\textbf{C.R.}\\\textbf{[\%] (\boldmath{$\uparrow$})}} & \shortstack[c]{\textbf{F1}\\\textbf{(\boldmath{$\uparrow$})}} & \shortstack[c]{\textbf{FPS}\\\textbf{(\boldmath{$\uparrow$})}} \\
    \midrule
    GTS & \cellcolor[RGB]{181,229,80}1.38 & \cellcolor[RGB]{181,229,80}2.80 & \cellcolor[RGB]{181,229,80}9.46 & \cellcolor[RGB]{181,229,80}18.53 & \cellcolor[RGB]{181,229,80}82.5 & 1.23 & \cellcolor[RGB]{230,92,90}1.91 & 10.25 & 21.43 & \cellcolor[RGB]{181,229,80}77.09 & 1.27 & 0.66 & 3.05 & 5.60 & \cellcolor[RGB]{75,139,190}71.30 & - & - \\
    \midrule
    MaskDINO & 1.81 & 4.40 & 16.76 & 35.65 & \cellcolor[RGB]{230,92,90}81.21 & 1.28 & 1.97 & \cellcolor[RGB]{230,92,90}9.16 & \cellcolor[RGB]{230,92,90}18.55 & \cellcolor[RGB]{75,139,190}76.63 & 1.26 & 0.68 & 3.08 & 5.58 & \cellcolor[RGB]{181,229,80}71.69 & 76.01 & \cellcolor[RGB]{181,229,80}5.95 \\
    OpenSeeD & \cellcolor[RGB]{230,92,90}1.64 & \cellcolor[RGB]{230,92,90}3.64 & \cellcolor[RGB]{230,92,90}13.8 & \cellcolor[RGB]{230,92,90}28.95 & 80.95 & \cellcolor[RGB]{75,139,190}1.15 & \cellcolor[RGB]{75,139,190}1.89 & \cellcolor[RGB]{75,139,190}8.91 & \cellcolor[RGB]{75,139,190}18.50 & \cellcolor[RGB]{230,92,90}74.39 & \cellcolor[RGB]{230,92,90}1.18 & \cellcolor[RGB]{75,139,190}0.653 & \cellcolor[RGB]{230,92,90}2.71 & \cellcolor[RGB]{230,92,90}4.78 & 66.5 & \cellcolor[RGB]{230,92,90}76.43 & \cellcolor[RGB]{230,92,90}1.43 \\
    OMG-Seg & \cellcolor[RGB]{75,139,190}1.54 & 4.67 & 16.46 & 36.00 & 76.01 & \cellcolor[RGB]{181,229,80}0.92 & 2.95 & 9.86 & 21.60 & 34.54 & \cellcolor[RGB]{75,139,190}1.14 & \cellcolor[RGB]{230,92,90}0.67 & \cellcolor[RGB]{181,229,80}2.62 & \cellcolor[RGB]{181,229,80}4.53 & 68.03 & \cellcolor[RGB]{75,139,190}77.99 & 1.33 \\
    UPPM (Ours) & 1.78 & \cellcolor[RGB]{75,139,190}3.30 & \cellcolor[RGB]{75,139,190}12.64 & \cellcolor[RGB]{75,139,190}25.53 & \cellcolor[RGB]{75,139,190}81.42 & \cellcolor[RGB]{230,92,90}1.19 & \cellcolor[RGB]{181,229,80}1.55 & \cellcolor[RGB]{181,229,80}5.65 & \cellcolor[RGB]{181,229,80}10.05 & 74.14 & \cellcolor[RGB]{181,229,80}1.1 & \cellcolor[RGB]{181,229,80}0.61 & \cellcolor[RGB]{75,139,190}2.64 & \cellcolor[RGB]{75,139,190}4.75 & \cellcolor[RGB]{230,92,90}70.76 & \cellcolor[RGB]{181,229,80}79.54 & \cellcolor[RGB]{75,139,190}2.28 \\
    \bottomrule
  \end{tabularx}
\end{adjustwidth}
\label{tab:combined-eval}
\end{table}

\begin{figure}[H]
  
  \includegraphics[width=0.85\linewidth]{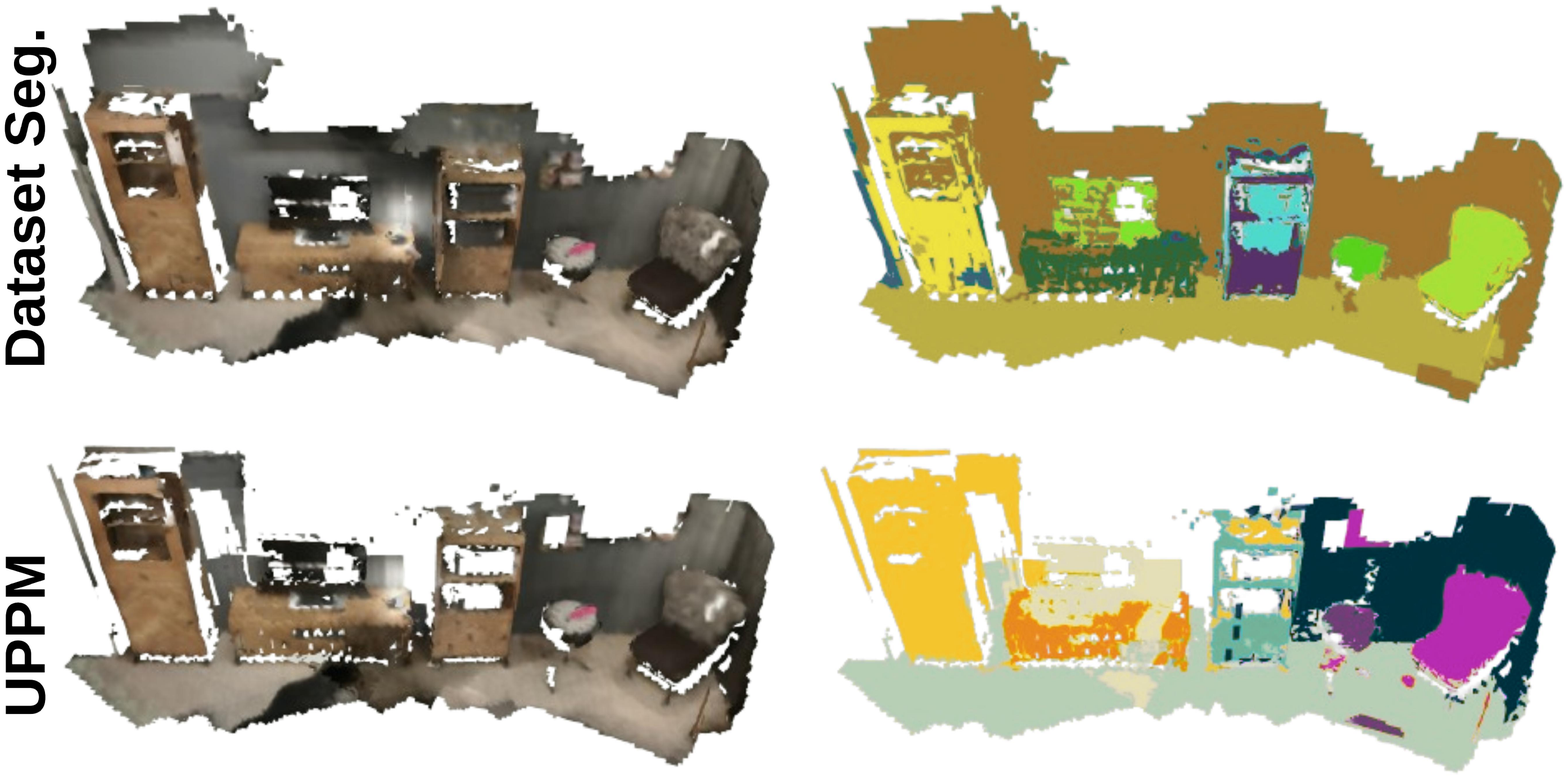}
  \caption{{Qualitative} 
 comparison on the RIO dataset~\cite{wald2019rio}. \textbf{Top:} Ground-truth panoptic segmentation and corresponding 3D map. \textbf{Bottom:} UPPM output with sharper object boundaries and more coherent semantic regions in cluttered, real-world~scenes.}
  \label{fig:rio-qualitative}
\end{figure}

To assess how thoroughly the reconstruction captures the ground truth scene, {Completeness} is defined as the mean absolute error from ground truth ($G$) to reconstruction ($R$).
UPPM attains outstanding completeness on Flat ($1.10$ cm) and competitive performance on ScanNet v2 ($1.78$ cm) and RIO (\mbox{$1.19$ cm}). This pattern reflects the influence of dataset characteristics: Flat~\cite{schmid2022panoptic}, being simulated, enables precise reconstruction and higher completeness, while the real-world conditions and noise in RIO~\cite{wald2019rio} and ScanNet v2~\cite{dai2017scannet} make achieving high completeness more challenging (\cref{fig:rio-qualitative}{). }
The metrics Accuracy and Completeness are asymmetrical, as~distances from ground truth points ($G$) to reconstructed map points ($R$) may differ from $R$ to $G$. These metrics compute distances by comparing each point in one set to its nearest neighbor in the other. A~larger Accuracy component suggests potential inaccuracies in reconstruction, while a larger Completeness component suggests potential~incompleteness.

To estimate how completely the reconstruction covers the ground truth, we use the {Completion Ratio}, defined as the percentage of observed ground truth points reconstructed within a $5$ cm threshold.
UPPM achieves strong completion ratios ($81.42\%$ on ScanNet v2, $74.14\%$ on RIO, $70.76\%$ on Flat). The~higher completion ratio on ScanNet v2 reflects the ability to leverage higher-quality data and more diverse semantic categories, while the lower ratio on RIO indicates the challenge of noisy real-world~data.

To quantify geometric similarity between the reconstructed and ground truth point clouds, {Chamfer distance} $d_{\text{ch}}(G,R)$ is used. This metric provides an overall assessment of how closely the reconstructed and ground truth point clouds match:
\begin{equation}
  d_{\text{ch}}(G,R) = \underbrace{\sum_{g \in G} \min_{r \in R} \left\| g - r \right\|^2_2}_{\text{Completeness}} +
  \underbrace{\sum_{r \in R} \min_{g \in G} \left\| g - r \right\|^2_2}_{\text{Accuracy}}
  \label{eq:chamfer}
\end{equation}
{UPPM} achieves exceptional Chamfer distance on RIO and competitive performance on ScanNet v2 and Flat (ScanNet v2: $12.64$ cm, RIO: $5.65$ cm, Flat: $2.64$ cm).
This lower Chamfer distance reflects the effectiveness of the multi-resolution Multi-TSDF approach, which balances accuracy and completeness. The~substantial improvements on RIO indicate UPPM's robustness to motion blur and sensor noise, while competitive performance on ScanNet v2 shows its ability to handle complex real-world scenes with diverse semantic~categories.

To estimate the worst-case reconstruction error, we use the {Hausdorff distance} ($d_H$), as~defined in \cref{eq:hausdorff}. This metric measures the greatest distance from a point in one set to the closest point in the other set:
\begin{equation}
  d_H(G,R) = \max \left( \max_{g \in G} \min_{r \in R} \|g - r\|_2, \max_{r \in R} \min_{g \in G} \|r - g\|_2 \right)
  \label{eq:hausdorff}
\end{equation}
{UPPM} achieves competitive Hausdorff distance performance with optimal performance on RIO (10.05\,cm vs. 18.55--21.60\,cm from competitors), demonstrating UPPM's robustness in minimizing worst-case reconstruction errors.
The Hausdorff distance on Flat ($4.75$ cm) highlights how the controlled environment helps minimize outlier errors. The~strong performance on ScanNet v2 ($25.53$ cm) demonstrates UPPM's ability to manage complex scenes with varied geometric structures while keeping worst-case errors within reasonable~limits.

To measure the ability to reconstruct fine-grained details while accounting for both false positives and false negatives, {F1-Score} is calculated after removing background classes (e.g., floor, ceiling, wall). UPPM achieves $2.28$ FPS on Flat (batch size 1), demonstrating computational efficiency suitable for real-time applications.
UPPM attains the highest F1-score ($79.54\%$), indicating better fine-detail reconstruction. This result is enabled by the custom NMS strategy, which removes duplicates while preserving details, and~by dynamic labeling that improves coverage without~over-segmentation.

\begin{figure}[H]
  
  \includegraphics[width=0.8\linewidth]{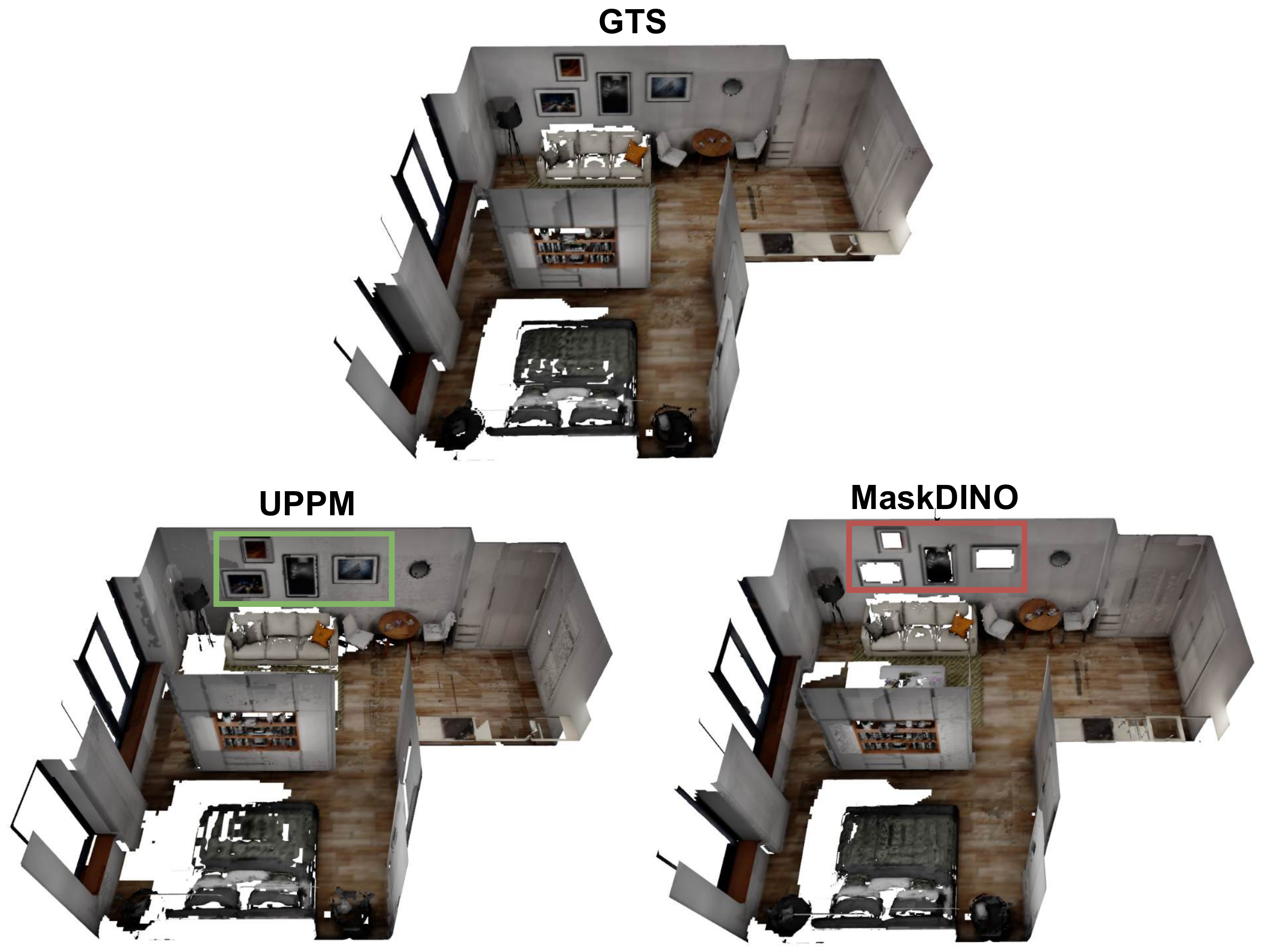}
  \caption{{Qualitative} 
 comparison of map reconstruction on the Flat dataset~\cite{schmid2022panoptic}. A~MaskDINO-based baseline attains high completion but misses fine-grained structures such as wall decorations, whereas UPPM recovers these details while preserving large-scale~geometry.}
  \label{fig:flatdataset-qualitative}
\end{figure}

The qualitative analysis reveals important insights about reconstruction quality. As~shown in \cref{fig:flatdataset-qualitative}, while MaskDINO achieves a better completion ratio on some datasets, it demonstrates better reconstruction for large structures like walls and floors but fails to capture intricate details such as paintings on walls, which hold significant semantic value. Conversely, UPPM successfully reconstructs these fine-grained elements, demonstrating its high capability in detailed scene~understanding.

\begin{figure}[H]

 \includegraphics[width=0.8\linewidth]{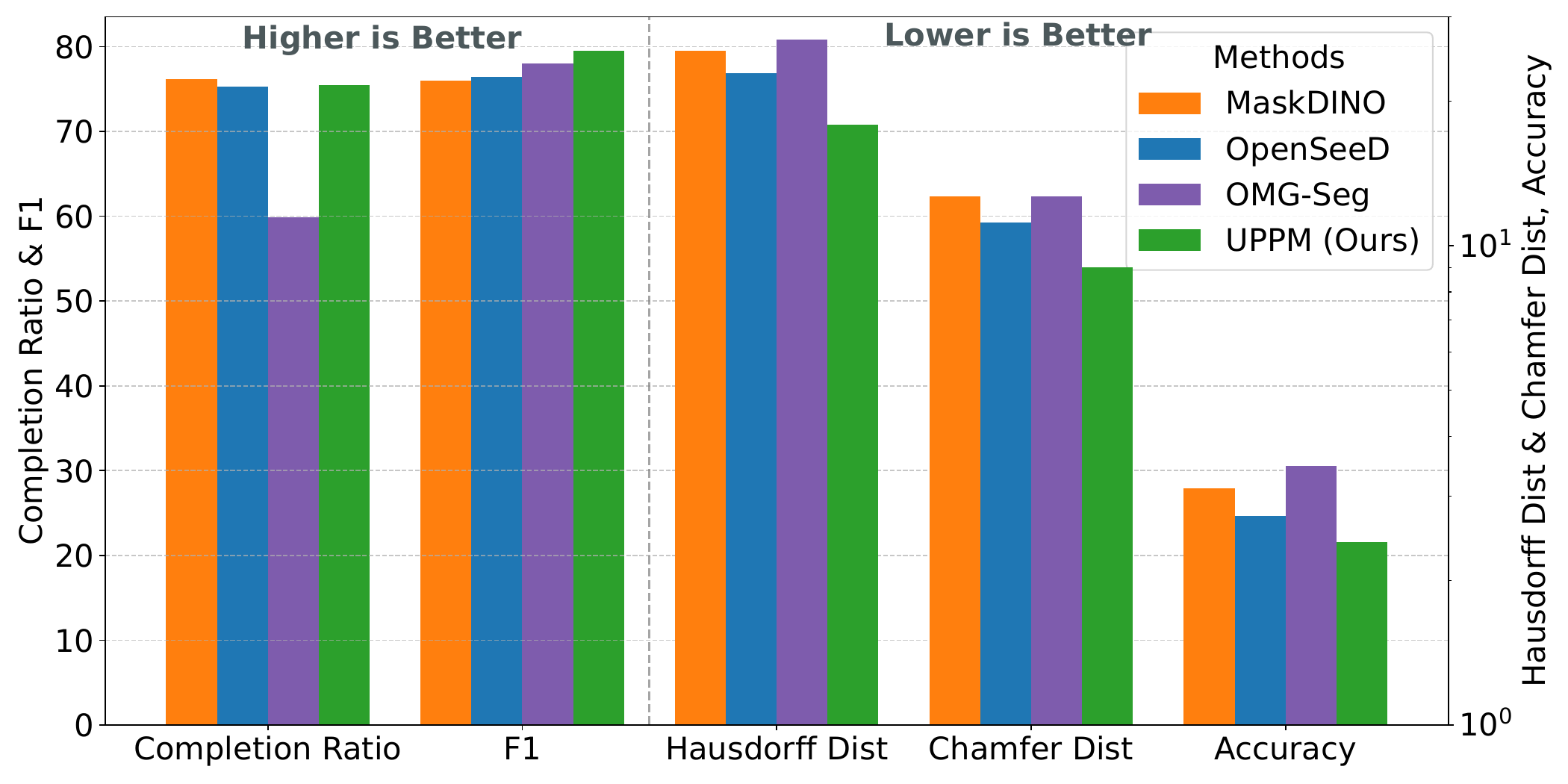}
 \caption{Aggregated performance across ScanNet v2, RIO, and~Flat. Completion Ratio and F1-score (left axis, higher is better) and Chamfer distance, Hausdorff distance, and~Accuracy (right axis, lower is better) indicate that UPPM offers a favorable trade-off between geometric fidelity and panoptic quality compared with alternative segmentation~backbones.}
 \label{fig:overall-performance}
\end{figure}

\clearpage
{Overall Performance Analysis: }
For comprehensive analysis across diverse scenarios, we present aggregated evaluation results for all datasets using a weighted average approach. As~shown {in} 
 \cref{fig:overall-performance}, UPPM shows robust performance across multiple metrics including Completion Ratio, F1-score, Chamfer distance, Hausdorff distance, and~Accuracy. The~Chamfer distance metric is more reliable and less sensitive to outliers than individual Accuracy and Hausdorff distance components, clearly demonstrating that UPPM outperforms other closed-set~\cite{li2023mask} and open-vocabulary~\cite{zhang2023simple, li2024omg} approaches in overall~performance.

\vspace{-10pt}

\vspace{-7pt}

\unskip
\subsection{Map-to-Map~Evaluation}
\label{sec:map-to-map-eval}
UPPM is compared against leading mapping pipelines including Kimera~\cite{rosinol2020kimera}, \mbox{PanMap~\cite{schmid2022panoptic}}, and~DHP~\cite{hu2024dhp}, representing state-of-the-art approaches in volumetric and semantic mapping.
The Chamfer-L1 metric provides robust assessment of overall reconstruction quality by measuring geometric fidelity with reduced sensitivity to outliers compared to L2~norm.

UPPM achieves remarkable performance across all geometric metrics (\cref{tab:geometry-comparison}), with~exceptional accuracy ($0.61$ cm), completeness ($1.10$ cm), and~Chamfer-L1 distance ($1.71$ cm). The~enhanced accuracy is enabled by the submap-based architecture that supports localized optimization, combined with semantic--geometric integration that improves object-level reconstructions. The~better completeness ($1.10$ cm vs. {$6.48$--$7.63$} 
 cm from competitors) demonstrates UPPM's ability to capture more complete scene geometry through its multi-resolution Multi-TSDF approach, while the outstanding Chamfer-L1 distance ($1.71$ cm vs. {$3.62$--$4.24$} cm) reflects the balanced optimization between accuracy and completeness. These results demonstrate UPPM's effectiveness in producing geometrically accurate maps while maintaining the benefits of dynamic semantic~labeling.

\begin{table}[H]

\caption{Geometric reconstruction quality on the Flat dataset. UPPM is compared against state-of-the-art methods, demonstrating exceptional performance in accuracy, completeness, and~Chamfer-L1~distance. {$\downarrow$ indicates lower is better.}}
\label{tab:geometry-comparison}
\setlength{\tabcolsep}{17.3pt}
\begin{tabular}{lccc}
\toprule
\textbf{Method} & \textbf{Acc.} \textbf{[cm]} \boldmath{$\downarrow$} & \textbf{Comp.} \textbf{[cm]} \boldmath{$\downarrow$} & \textbf{Chamfer-L1} \textbf{[cm]} \boldmath{$\downarrow$} \\
\midrule
Kimera & 0.76 & 6.48 & 3.62 \\
Panmap & 0.86 & 7.63 & 4.24 \\
DHP & 0.73 & 6.58 & 3.65 \\
UPPM (Ours) & {0.61} 
 & {1.1} & {1.71} \\
\bottomrule
\end{tabular}
\end{table}
\unskip

\subsection{Segmentation-to-Segmentation~Evaluation}
\label{sec:seg-to-seg-eval}

As shown in \cref{tab:segmentation-eval}, UPPM achieves the highest {Panoptic Quality} (PQ) ($0.414$), {Recognition Quality} (RQ) ($0.475$), and~{mean Average Precision} (mAP) across multiple IoU thresholds, with~competitive {Segmentation Quality} (SQ) ($0.845$). The~better PQ and RQ performance stems from the unified semantics approach that reduces label ambiguity and improves instance recognition, while the competitive SQ reflects the precision of instance masks guided by dynamic descriptors and enhanced duplicate suppression through custom NMS. The~highest mAP scores across all IoU thresholds ($0.549$, $0.521$, $0.475$) demonstrate UPPM's consistent performance in object detection and semantic retrieval, outperforming MaskDINO ($0.546$, $0.516$, $0.470$) and other competitors. These results validate the effectiveness of the dynamic labeling approach in producing high-quality panoptic segmentations that maintain both semantic consistency and spatial~accuracy.

\begin{table}[H]

\setlength{\tabcolsep}{7.2pt}
  \renewcommand{\arraystretch}{1.2}
\renewcommand{\aboverulesep}{.1pt}
\renewcommand{\belowrulesep}{.1pt}

\caption{Panoptic segmentation quality on the Flat dataset. UPPM is evaluated against other methods on key metrics, with~highlight colors indicating performance ranking: \textcolor[RGB]{181,229,80}{best}, \textcolor[RGB]{75,139,190}{second best}, and \textcolor[RGB]{230,92,90}{\mbox{third best}}. {$\uparrow$ indicates higher is better.}} 
\begin{tabularx}{\textwidth}{lcccccc}

\toprule
\textbf{Method} & \textbf{PQ} \boldmath{$\uparrow$} & \textbf{RQ} \boldmath{$\uparrow$} & \textbf{SQ} \boldmath{$\uparrow$} & \textbf{mAP-(0.3)} \boldmath{$\uparrow$} & \textbf{mAP-(0.4)} \boldmath{$\uparrow$} & \textbf{mAP-(0.5)} \boldmath{$\uparrow$} \\
\midrule
MaskDINO & \cellcolor[RGB]{75,139,190}0.406 & \cellcolor[RGB]{75,139,190}0.470 & \cellcolor[RGB]{181,229,80}0.851 & \cellcolor[RGB]{75,139,190}0.546 & \cellcolor[RGB]{75,139,190}0.516 & \cellcolor[RGB]{75,139,190}0.470 \\
OMG-Seg & 0.164 & 0.200 & 0.498 & 0.287 & 0.244 & 0.200 \\
Detectron2 & \cellcolor[RGB]{230,92,90}0.343 & \cellcolor[RGB]{230,92,90}0.432 & \cellcolor[RGB]{230,92,90}0.787 & \cellcolor[RGB]{230,92,90}0.499 & \cellcolor[RGB]{230,92,90}0.473 & \cellcolor[RGB]{230,92,90}0.432 \\
UPPM (Ours) & \cellcolor[RGB]{181,229,80}0.414 & \cellcolor[RGB]{181,229,80}0.475 & \cellcolor[RGB]{75,139,190}0.845 & \cellcolor[RGB]{181,229,80}0.549 & \cellcolor[RGB]{181,229,80}0.521 & \cellcolor[RGB]{181,229,80}0.475 \\
\bottomrule
\end{tabularx}
\label{tab:segmentation-eval}
\end{table}
\unskip

{\subsection{Computational Overhead~Analysis}}
\label{sec:computational-overhead}
{To address the computational requirements of UPPM's vision--language pipeline, we provide a detailed timing breakdown in} \cref{tab:timing-breakdown}.
{The experiments were conducted on an NVIDIA A100 (80~GB) GPU with 256~GB of RAM.}

\begin{table}[H]
\setlength{\tabcolsep}{19pt}

\caption{{Per-frame timing breakdown of UPPM components on the Flat dataset. Times are averaged over 762 frames at {640} 
 $\times$ 480 resolution.}}
\label{tab:timing-breakdown}
\begin{tabularx}{\textwidth}{lcc}

\toprule
{\textbf{Component}} & {\textbf{Time (ms)}} & {\textbf{\% of Total}} \\
\midrule
{VLFE++ (Tag2Text)} & {105} & {24.0\%} \\
{Semantic Retrieval (FAISS + MPNet)} & {8} & {1.8\%} \\
{Grounding-DINO (Swin-T)} & {120} & {27.4\%} \\
{SAM (ViT-H)} & {180} & {41.1\%} \\
{Custom NMS} & {2} & {0.5\%} \\
{Unified Panoptic Fusion} & {22.8} & {5.2\%} \\
\midrule
{{Total} 
} & {{437.8}} & {{100\%}} \\
\bottomrule
\end{tabularx}
\end{table}

{UPPM achieves 2.28 FPS on the Flat dataset, which is slower than the closed-set MaskDINO baseline (5.95 FPS) due to the additional vision--language processing.}
{However, this overhead is justified by the open-vocabulary capability and improved panoptic quality.}
{Compared to PanMap with MaskDINO (5.95 FPS), UPPM trades 62\% of the frame rate for: (i) open-vocabulary recognition without retraining, (ii) 3.5\% improvement in F1-score, and~(iii) language-conditioned queryability.}

\subsection{Ablation~Studies}
\label{sec:ablation_studies}

The objective of the ablation studies is to gain a deeper understanding of the contributions made by different components in~UPPM.

\subsection{Unified~Semantics}
\label{sec:unified-semantics}
\begin{figure}[H]
  
  \includegraphics[width=0.8\linewidth]{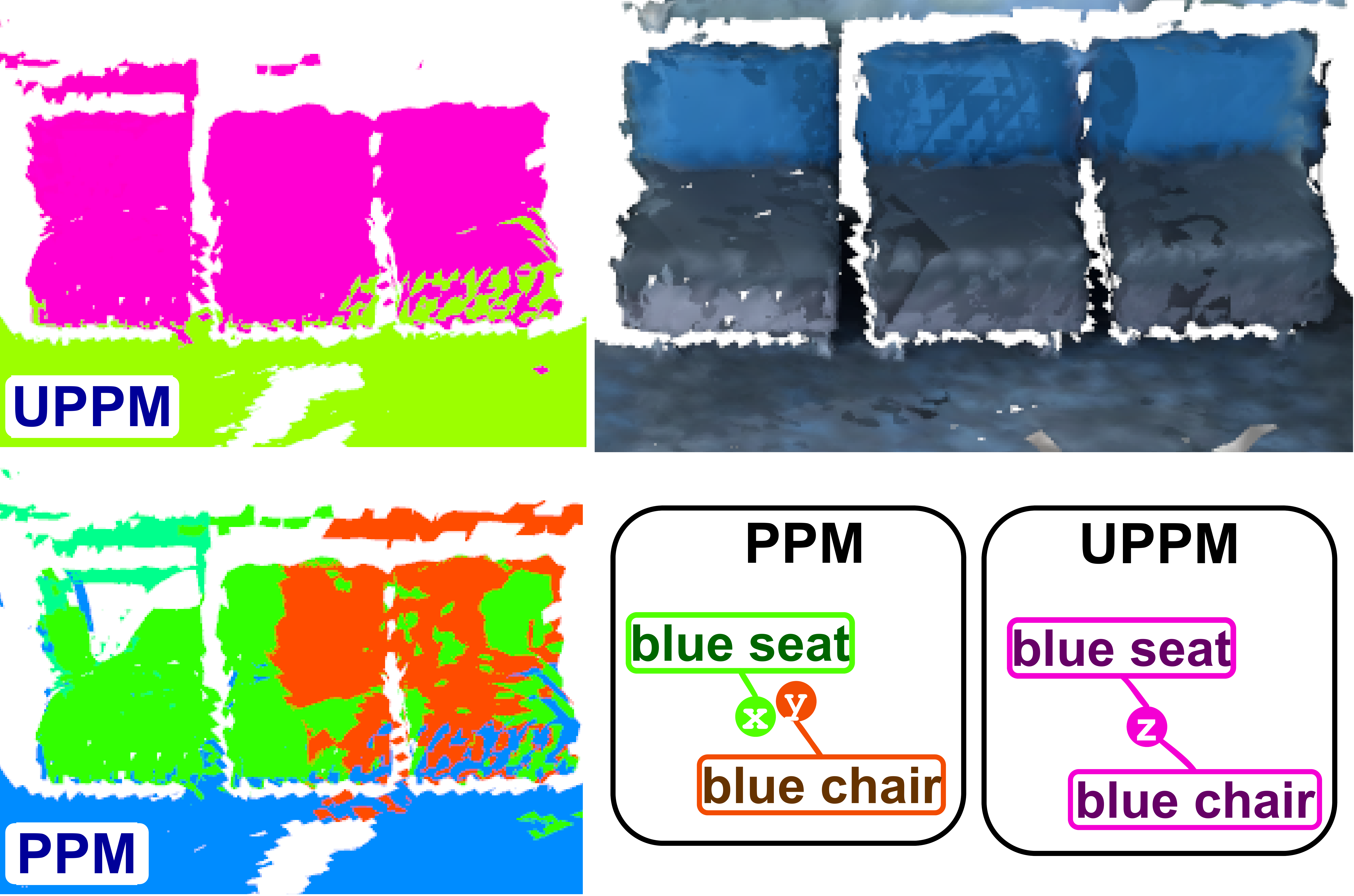}
  \caption{{Effect} 
 of unified semantics on 3D panoptic segmentation. PPM, which disables unified semantics, assigns multiple ambiguous labels ($x, y, z$) to the same object, whereas UPPM produces a single consistent semantic class while retaining rich open-vocabulary~descriptors.}
  \label{fig:scannet-unified-semantics-qualitative}
\end{figure}

In this study, we examine the effects of deactivating the unified semantic mechanism, which we call PPM (Promptable Panoptic Mapping), to~assess the impact of unified semantics on map reconstruction. As~shown in \cref{tab:ablation}, UPPM outperforms PPM across all metrics on the Flat dataset, achieving better completion, accuracy, and~Chamfer distance, in~addition to the added benefit of unified semantics. As~shown in \cref{fig:scannet-unified-semantics-qualitative}, PPM lacks dynamically labeled classes, leading to ambiguity when many semantic categories attach to the same object. In~comparison, UPPM exhibits consistent behavior by reliably assigning semantic classes to objects while maintaining richness in dynamic~labeling.

\unskip

\begin{table}[H]
\small
  \setlength{\tabcolsep}{28pt}
  \caption{Results of quantitative ablation experiments on Flat dataset. Abbreviations: Comp. = Completion, Acc. = Accuracy, Chamf. = Chamfer~distance. {$\downarrow$ indicates lower is better.}} 
  \begin{tabularx}{\textwidth}{lccc}
    \toprule
     \multirow{2}{*}{\textbf{Method}} & \textbf{Comp.} & \textbf{Acc.} & \textbf{Chamf.} \\
    & \textbf{[cm] ($\downarrow$)} & \textbf{[cm] ($\downarrow$)} & \textbf{[cm] ($\downarrow$)} \\
    \midrule
    UPPM & {1.1} & {0.61} & {2.64} \\
    UPPM with tags & 1.2 & 0.63 & 2.77 \\
    \midrule
    PPM & 1.18 & 0.65 & 2.79 \\
    PPM with tags & 1.22 & 0.66 & 2.92 \\
    \bottomrule
  \end{tabularx}
  \label{tab:ablation}
\end{table}
\unskip

\subsubsection{Non-Maximum~Suppression}
\label{sec:nms}
We compare traditional NMS with our custom NMS on {Flat} 
 (\cref{tab:custom-nms-vs-traditional-nms,tab:nms-comparison}).

Custom NMS improves the completion ratio by 8.27\% (62.11\% $\to$ 70.38\%) at unchanged Chamfer (2.79\,cm), widening reconstructed coverage by removing overlaps and redundancies. The~8.27\% gain arises from context-aware suppression that prioritizes caption-derived labels and removes duplicates across semantically equivalent detections. This improvement demonstrates that our custom NMS strategy effectively balances duplicate removal with coverage preservation, achieving better completion without sacrificing geometric~accuracy.

{To further quantify the impact on panoptic segmentation quality, we conducted an ablation study on the challenging RIO dataset (sequence 0cac7578) comparing UPPM with and without NMS (\cref{tab:nms-pq-ablation}). The~results demonstrate substantial improvements across all panoptic quality metrics: PQ increases by 63.5\% (17.62\% $\to$ 28.82\%), RQ improves by 58.8\% (23.44\% $\to$ 37.23\%), and~SQ gains 3.0\% (75.16\% $\to$ 77.39\%). The~near-doubling of true positives (+99.6\%) alongside a 16.9\% reduction in false negatives confirms that coordinate-based NMS effectively prevents instance fragmentation while improving recognition quality.
The absolute PQ values reflect the inherent difficulty of open-vocabulary panoptic segmentation on RIO, which exhibits significant motion blur and sensor noise; the key observation is the substantial relative improvement enabled by coordinate-based NMS.}

\begin{table}[H]
\small
  
  \caption{Quantitative NMS results on the Flat dataset. The~custom NMS significantly improves the completion ratio while maintaining the same Chamfer~distance. {$\downarrow$ indicates lower is better; $\uparrow$ indicates higher is better.}} 
      \setlength{\tabcolsep}{15pt}
  \begin{tabularx}{\textwidth}{lcc}
    \toprule
    \textbf{Method} & \textbf{Chamfer Dist. [cm] (\boldmath{$\downarrow$})} & \textbf{Comp. Ratio [$<$5cm \%] (\boldmath{$\uparrow$})} \\
    \midrule
    PPM & 2.79 & {70.38} \\
    PPM w/o NMS & 2.79 & 62.11 \\
    \midrule
    Enhancement & 0.00 & {+8.27\%} \\
    \bottomrule
  \end{tabularx}
  \label{tab:nms-comparison}
\end{table}

\vspace{-10pt}

\begin{table}[H]
\small
  
  \caption{{NMS ablation study on panoptic quality metrics (RIO dataset). Custom NMS substantially improves PQ, SQ, and~RQ by reducing instance fragmentation. {$\downarrow$ indicates lower is better; $\uparrow$ indicates higher is better.}}} 
     \setlength{\tabcolsep}{10.35pt}
  \begin{tabularx}{\textwidth}{lccccc}
    \toprule
    \textbf{Method} & \textbf{PQ [\%] (\boldmath{$\uparrow$})} & \textbf{SQ [\%] (\boldmath{$\uparrow$})} & \textbf{RQ [\%] (\boldmath{$\uparrow$})} & \textbf{TP (\boldmath{$\uparrow$})} & \textbf{FN (\boldmath{$\downarrow$})} \\
    \midrule
    {UPPM w/o NMS} & {17.62} & {75.16} & {23.44} & {283} & {1669} \\
    {UPPM (Ours)} & {{28.82}} & {{77.39}} & {{37.23}} & {{565}} & {{1387}} \\
    \midrule
    {Improvement} & {+63.5\%} & {+3.0\%} & {+58.8\%} & {+99.6\%} & {$-$16.9\%} \\
    \bottomrule
  \end{tabularx}
  \label{tab:nms-pq-ablation}
\end{table}

\subsubsection{Blurry Frame~Filtering}
\label{sec:blurry-frames-filtering}
We skip frames flagged as blurry by the caption model (\cref{fig:system-overview}). This filtering strategy has a negligible effect on Flat, a small impact on ScanNet v2 (0.84\% flagged), and~a larger effect on RIO (3\% flagged; manual inspection suggests $>21\%$ affected).
On RIO, UPPM reduces Chamfer distance by 16.675\% and increases completion ratio by 6.47\%. These gains arise because filtering out low-quality frames reduces geometric inconsistencies and error accumulation during~fusion.

\subsubsection{Identified~Tags}
As shown in \cref{fig:system-overview}, we offer the option to use the identified tags in the pipeline. For~the Flat dataset, as~shown in \cref{tab:ablation}, integrating the identified tags to both PPM and UPPM results in marginal improvement in the completion ratio and all other metrics in the case of UPPM. This demonstrates that semantic information adds to more comprehensive scene~knowledge.

\section{Conclusions}
UPPM integrates open-vocabulary vision--language perception with panoptic 3D mapping through a simple dynamic descriptor design. Each object's dynamic descriptor collects rich open-vocabulary labels, maps them to a unified category and size prior, and~fuses the result into a multi-resolution multi-TSDF map. This aggregation reconciles the long-tailed semantics of modern foundation models with the consistency requirements of panoptic mapping: it preserves open-vocabulary expressiveness while preventing synonyms from fracturing instances across frames. Experiments on ScanNet v2, RIO, and Flat show that UPPM delivers the best reconstruction accuracy and panoptic segmentation quality. It outperforms strong closed-set and open-set baselines, and it~enables downstream language-conditioned tasks such as object retrieval. Ablation studies attribute these gains to the unified semantics and dynamic descriptors, with~custom non-maximum suppression and blurry-frame filtering further improving completeness without sacrificing~accuracy.

Apart from the numerous advantages of the proposed method, UPPM inherits limitations from its vision-language backbones. The~quality of the captions and detections bounds the accuracy of the dynamic descriptors, and~residual semantic drift can persist in highly cluttered or long-tailed settings. The~system also incurs non-trivial computational overhead from high-capacity language models and high-resolution map updates. Future work should explore tighter coupling between semantics and geometry, lighter-weight labeling models for real-time deployment, improved temporal reasoning for long-term and multi-session mapping, and~extensions to dynamic or outdoor~environments.

\vspace{+6pt}

\authorcontributions{Conceptualization, M.A.M., S.Z. and G.F.; methodology, M.A.M.; software, M.A.M. and R.S.; validation, M.A.M. and R.S.; formal analysis, M.A.M.; investigation, M.A.M.; resources, G.F.; data curation, M.A.M.; writing---original draft preparation, M.A.M.; writing---review and editing, R.S., G.K., S.Z. and G.F.; visualization, M.A.M. and R.S.; supervision, S.Z. and G.F. All authors have read and agreed to the published version of the manuscript.}

\funding{The work was supported by the grant for research centers in the field of AI provided by the Ministry of Economic Development of the Russian Federation in accordance with the agreement 000000C313925P4F0002 and the agreement with №139-10-2025-033.}

\institutionalreview{Not applicable.}

\informedconsent{Not applicable.}

\dataavailability{{The source code is publicly available at} \url{https://Unified-Promptable-Panoptic-Mapping.github.io} {(accessed on 25 January 2026).}
}

\conflictsofinterest{{Authors Mohamad Al Mdfaa, Raghad Salameh and Gonzalo Ferrer are employed by the company Applied AI Institute. The remaining authors declare that the research was conducted in the absence of any commercial or financial relationships that could be construed as a potential conflict of interest.}} 

\begin{adjustwidth}{-\extralength}{0cm}
\reftitle{References}

\PublishersNote{}

\end{adjustwidth}

\end{document}